\documentclass[twoside]{article}

\usepackage[accepted]{aistats2025}

\usepackage[utf8]{inputenc} 
\usepackage[T1]{fontenc}    
\usepackage{url}            
\usepackage{booktabs}       
\usepackage{amsfonts}       
\usepackage{nicefrac}       
\usepackage{microtype}      
\usepackage{xcolor}         
\usepackage[pdftex]{graphicx}
\usepackage{amsmath, amsthm, amssymb}
\usepackage{comment}
\usepackage{subfig}
\usepackage{booktabs}
\usepackage{multirow}
\usepackage{caption}
\usepackage{algorithm}
\usepackage{algpseudocode}
\usepackage{arydshln}
\usepackage[toc,page]{appendix}
\usepackage{mathtools}
\usepackage{caption} 
\usepackage{bbm}
\usepackage[colorlinks,citecolor=blue]{hyperref}

\newtheorem{theorem}{Theorem}
\newtheorem{assumption}{Assumption}
\newtheorem{lemma}{Lemma}
\newtheorem{definition}{Definition}

\newcommand{\std}[1]{\footnotesize{\color{gray}$\pm$#1}}
\definecolor{LightGreen}{rgb}{0.55,0.9,0.55}
\definecolor{MediumGreen}{rgb}{0.3,0.85,0.3}
\definecolor{DarkGreen}{rgb}{0.,0.7,0.}
\definecolor{LightRed}{rgb}{1,0.6,0.6}
\definecolor{MediumRed}{rgb}{0.9,0.4,0.4}
\definecolor{DarkRed}{rgb}{0.85,0.1,0.1}
\usepackage[round]{natbib}

\begin{document}

%

%

\twocolumn[

\aistatstitle{Conformal Prediction for Federated Graph Neural Networks with Missing Neighbor Information}

\aistatsauthor{ Ömer Faruk Akgül \And 
Rajgopal Kannan \And  Viktor Prasanna }

\aistatsaddress{ University of Southern California \And  DEVCOM Army Research Office \And University of Southern California } ]

\begin{abstract}
  Graphs play a crucial role in data mining and machine learning, representing real-world objects and interactions. As graph datasets grow, managing large, decentralized subgraphs becomes essential, particularly within federated learning frameworks. These frameworks face significant challenges, including missing neighbor information, which can compromise model reliability in safety-critical settings. Deployment of federated learning models trained in such settings necessitates quantifying the uncertainty of the models. This study extends the applicability of Conformal Prediction (CP), a well-established method for uncertainty quantification, to federated graph learning. We specifically tackle the missing links issue in distributed subgraphs to minimize its adverse effects on CP set sizes. We discuss data dependencies across the distributed subgraphs and establish conditions for CP validity and precise test-time coverage. We introduce a Variational Autoencoder-based approach for reconstructing missing neighbors to mitigate the negative impact of missing data. Empirical evaluations on real-world datasets demonstrate the efficacy of our approach, yielding smaller prediction sets while ensuring coverage guarantees. 
\end{abstract}

\section{Introduction} \label{sec:intro}

Graph Neural Networks (GNNs) have significantly advanced graph data mining, demonstrating strong performance across various domains, including social platforms, e-commerce, transportation, bioinformatics, and healthcare \citep{hamilton2018inductiverepresentationlearninglarge,  kipf2017semisupervisedclassificationgraphconvolutional, wu2022graph, zhang2021graph}. In many real-world scenarios, graph data is inherently distributed due to the nature of data generation and collection processes \citep{zhou2020graph}. For example, data from social networks, healthcare systems, and financial institutions \citep{liu2019geniepath} is often generated by multiple independent entities, leading to fragmented and distributed graph structures. This distributed nature of graph data poses unique challenges when training GNNs, such as the need to address data privacy, ownership, and regulatory constraints \citep{zhang2021subgraph}.

Federated Learning (FL) emerges as a solution, allowing collaborative model training without centralized data sharing \citep{mcmahan2017communication, kairouz2021advances}. FL addresses data isolation issues and has been widely used in various applications, including computer vision and natural language processing \citep{li2020review}. However, applying FL to graph data introduces unique challenges, such as incomplete node neighborhoods and missing links across distributed subgraphs \citep{zhang2021subgraph}. These missing connections can degrade model performance and increase uncertainty, underscoring the need for robust uncertainty quantification techniques.

\begin{figure*}[htbp]
    \centering
        \includegraphics[width=0.82\linewidth]{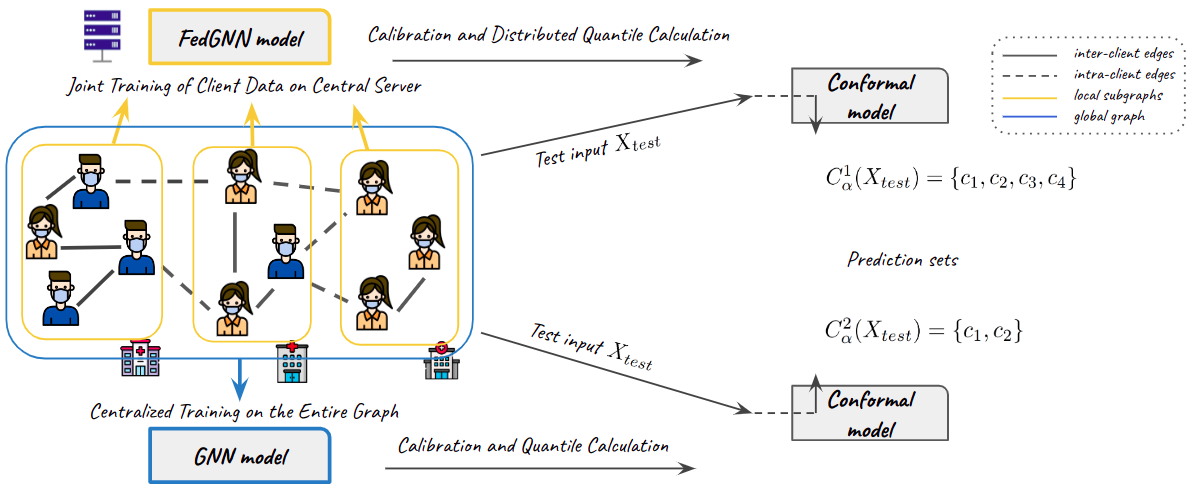}
    \caption{\textbf{Overview of federated conformal prediction for graph-structured data.} A scenario involving patient data shared across three hospitals. It distinguishes between intra-client (solid lines) and inter-client (dashed lines) interactions, with the former stored in hospital databases and the latter often missing in federated settings despite their real-life presence. The FedGNN model leverages a federated GNN to optimize a global model through local client updates. In contrast, the conventional GNN model is trained under an ideal scenario where all connections (both solid and dashed) are accessible, providing a benchmark for comparison. The figure also highlights how missing inter-client links contribute to inefficiencies in the conformal prediction set size, as demonstrated in the prediction sets \(C^1_{\alpha}(X_{\text{test}})\) and \(C^2_{\alpha}(X_{\text{test}})\).}
    \label{fig-overview}
\end{figure*}

Conformal Prediction \citep{vovk2005algorithmic} offers a promising framework for producing statistically guaranteed uncertainty estimates, providing user-specified confidence levels to construct prediction sets with provable coverage guarantees. Specifically, with a miscoverage level \( \alpha \in (0, 1) \), CP uses calibration data to generate prediction sets for new instances, ensuring the true outcome is contained within them with probability at least \( 1 - \alpha \).

While CP has been explored in natural language processing \citep{kumar2023conformal}, computer vision \citep{angelopoulos2020uncertainty}, federated learning \citep{lu2023federated}, and GNNs \citep{zargarbashi2023conformal, huang2024uncertainty}, its application in federated graph learning remains underexplored. A primary challenge is ensuring the \emph{exchangeability} assumption—critical for CP's validity—holds in partitioned graph data, which may not be the case due to data heterogeneity across clients.

In this paper, we investigate conformal prediction within a federated graph learning framework, where multiple clients, each with distinct local data distributions \(P_k\) and experiencing missing neighbor information, collaboratively train a shared global model. Our goal is to establish marginal coverage guarantees for prediction sets derived from unseen data sampled from a global distribution \(Q_{\text{test}}\), which is a mixture of the clients' individual distributions \(P_k\) weighted by coefficients \(p_k\). The heterogeneity among client distributions can violate the exchangeability assumption, compromising coverage guarantees and leading to larger, less efficient prediction sets \citep{huang2024uncertainty}. Additionally, missing neighbor information exacerbates uncertainty by limiting dataset completeness.

We extend the theoretical framework of {\it partial exchangeability} to graphs within the federated learning setting, addressing the challenges posed by data heterogeneity across the client subgraphs. Our analysis reveals inefficiencies in the size of the conformal prediction sets attributable to missing links. To counteract these inefficiencies, we introduce a novel framework designed to generate missing links across clients, thereby optimizing the size of CP sets.

Our main contributions are summarized as follows:

\begin{itemize}
\item We extend Conformal Prediction to federated graph settings, establish the necessary conditions for CP validity and derive theoretical statistical guarantees.
\item We identify the impact of missing links on prediction set sizes due to missing links and propose a method for local subgraph completion to enhance CP efficiency.
\item We demonstrate the effectiveness of our approach through empirical evaluation on four benchmark datasets, showing improved efficiency of CP in federated graph scenarios.
\end{itemize}
\section{Related Work}

Recent advancements have applied Conformal Prediction to graph machine learning to enhance uncertainty quantification in (GNNs). \citep{clarkson2023distribution} improved calibration and prediction sets for node classification in inductive learning scenarios. \citep{huang2024uncertainty} and \citep{zargarbashi2023conformal} focused on reducing prediction set sizes, with the latter enhancing efficiency through the diffusion of node-wise conformity scores and leveraging network homophily.

In Federated Learning, \citep{lu2023federated} extended CP techniques to address data heterogeneity, providing theoretical guarantees for uncertainty quantification in distributed settings. \citep{zhang2021subgraph} introduced FedSage and FedSage+, methods for training graph mining models on distributed subgraphs, tackling data heterogeneity and missing links. \citep{baek2023personalized} explored personalized weight aggregation based on subgraph similarity in a personalized subgraph FL framework. \citep{tan2022fedproto} proposed FedProto, which constructs prototypes from local client data to enhance learning across subgraphs, though it does not address privacy concerns related to sharing prototypes.

Despite these efforts, existing work does not fully address the challenges of missing links and subgraph heterogeneity in graph conformal settings. Our work is the first to propose a CP method specifically designed for federated graph learning, addressing both exchangeability violations and inefficiencies caused by missing neighbor information.
\section{Preliminaries} \label{sec:background}

\subsection{Conformal Prediction}

Conformal prediction is a framework for uncertainty quantification that provides rigorous statistical guarantees. We focus on the split conformal prediction method \citep{vovk2005algorithmic}, notable for its computational efficiency. The method defines a non-conformity measure \( S: \mathcal{X} \times \mathcal{Y} \rightarrow \mathbb{R} \), which quantifies how atypical the true label \( y \) is for the input \( x \) according to the model's predictions. For classification tasks, \( S(x, y) \) might be defined as \( 1 - f_y(x) \), where \( f_y(x) \) is the estimated probability of class \( y \) given \( x \).

\subsubsection{Quantile Calculation and Prediction Set Construction}

Using a calibration dataset \( \mathcal{D}_{\text{calib}} = \{(x_i, y_i)\}_{i=1}^n \), we compute the non-conformity scores \( S_i = S(x_i, y_i) \) for each calibration example. The cutoff value \( \hat{q} \) is then determined as the \( (1 - \alpha)(1 + \frac{1}{n}) \)-th empirical quantile of these scores, i.e.,
\( \hat{q} = \text{quantile}\left( \{S_1, \ldots, S_n\},\, (1 - \alpha)\left(1 + \frac{1}{n}\right) \right) \). Given a new input \( x \), the prediction set is constructed as \( C_\alpha(x) = \{ y \in \mathcal{Y} : S(x, y) \leq \hat{q} \} \). Under the assumption of exchangeability of the data, this method guarantees that the true label \( y \) will be contained in \( C_\alpha(x) \) with probability at least \( 1 - \alpha \).

Adaptive Prediction Sets (APS)\citep{romano2020classification} construct prediction sets by accumulating class probabilities. Given a probabilistic classifier that outputs estimated class probabilities \( f(x) = (f_1(x), \ldots, f_{|\mathcal{Y}|}(x)) \), where \( f_j(x) \) is the estimated probability of class \( j \) for input \( x \), we sort the classes in descending order to obtain a permutation \( \pi \) such that \( f_{\pi(1)}(x) \geq f_{\pi(2)}(x) \geq \ldots \geq f_{\pi(|\mathcal{Y}|)}(x) \). The cumulative probability up to the \( k \)-th class is \( V(x, k) = \sum_{j=1}^k f_{\pi(j)}(x) \). For each calibration example \( (x_i, y_i) \), we compute the non-conformity score \( S_i = V(x_i, k_i) \), where \( k_i \) is the rank of the true label \( y_i \) in the sorted class probabilities for \( x_i \). The cutoff value \( \hat{q} \) is then determined as before. The prediction set \( C_\alpha(x) \) includes the top \( k^* \) classes, where \( k^* = \min \left\{ k : V(x, k) \geq \hat{q} \right\} \) and \( C_\alpha(x) = \{ \pi(1), \ldots, \pi(k^*) \} \).

\subsubsection{Evaluation Metrics}

Our goal is to achieve valid marginal coverage while minimizing the size of the prediction sets. The inefficiency is measured as
\( \text{Inefficiency}_\alpha = \frac{1}{m} \sum_{j=1}^m |C_\alpha(x_j)| \),
where \( \mathcal{D}_{\text{test}} = \{ (x_j, y_j) \}_{j=1}^m \) is the test set. Empirical coverage is calculated as
\( \text{Coverage}_\alpha = \frac{1}{m} \sum_{j=1}^m \mathbbm{1}\{ y_j \in C_\alpha(x_j) \} \),
representing the proportion of test examples where the true label is included in the prediction set.

\subsection{GNNs and Federated Graph Learning}

Graph Neural Networks effectively capture structural information and node features in graph-structured data \citep{kipf2016semi}. Consider a graph $\mathcal{G} = (\mathcal{V}, \mathcal{E})$, where $\mathcal{V}$ is the set of $n$ nodes and $\mathcal{E}$ is the set of edges. Each node \( v \in \mathcal{V} \) has a feature vector \( x_v \in \mathbb{R}^d \), forming the matrix \( X = \{x_v\}_{v \in \mathcal{V}} \in \mathbb{R}^{n \times d} \).

In node classification, we predict labels based on node features and the graph structure, following a transductive learning setting where \( \mathcal{G} \) is available during training and testing, but test labels are withheld. In the conformal prediction setting, we introduce an additional subset called the calibration set, alongside the traditional subsets in the transductive setting, partitioning \( \mathcal{V} \) into training, validation, calibration, and test subsets: \( \mathcal{V}_{\text{train}} \), \( \mathcal{V}_{\text{valid}} \), \( \mathcal{V}_{\text{cal}} \), and \( \mathcal{V}_{\text{test}} \).

Federated Graph Neural Networks extend GNNs to federated learning scenarios with data distributed across multiple clients \citep{wu2021fedgnn}. A central server coordinates with \( K \) clients, each holding a subgraph of \( \mathcal{G} \). We use Federated Averaging (FedAvg) \citep{mcmahan2017communication} to aggregate client model updates, updating the global model \( \theta \) by computing a weighted average:
\( \theta = \sum_{k=1}^K \frac{n_k}{n} \theta_k \),
where \( \theta_k \) are local model parameters, \( n_k \) is the number of samples at client \( k \), and \( n = \sum_{k=1}^K n_k \). This approach preserves data privacy by allowing clients to train locally while contributing to the global model without sharing raw data \citep{mcmahan2017communication}.

\subsection{Variational Autoencoders} \label{sec:vae}

Variational Autoencoders (VAEs) \citep{kingma2013auto} and their extension to graph data, Variational Graph Autoencoders (VGAEs) \citep{kipf2016variational}, are fundamental to our approach for generating node features and predicting edges. Both methods utilize deep learning and Bayesian inference to learn latent representations by optimizing the evidence lower bound (ELBO), balancing reconstruction loss and the Kullback-Leibler (KL) divergence between the approximate and prior distributions. The ELBO is defined as: 
$\mathcal{L}(\theta, \phi; x) = \mathbb{E}_{q_\phi(z|x)}[\log p_\theta(x|z)] - D_{\text{KL}}[q_\phi(z|x) \| p(z)]$, 
where $q_\phi(z|x)$ approximates the latent variable $z$, and $p(z)$ is the prior. In our model, VAEs generate node features, while VGAEs extend this framework to graph data, learning latent representations of graph structures for tasks such as edge prediction.
\section{Challenges of Conformal Prediction on Federated Graphs} \label{sec:challenges}

Conformal Prediction on federated graphs faces several challenges that need to be addressed to ensure its applicability and effectiveness in real-world applications. In this section, we elaborate on these challenges. In Section \ref{sec:method}, we discuss how we address them.

\textit{Exchangeability:} A significant challenge in federated graph CP is the violation of the exchangeability principle, which traditional CP methods rely upon \citep{vovk2005algorithmic}. Consider a federated graph learning setting where nodes of the overall graph \( \mathcal{V} \) are partitioned into training, validation, calibration, and test sets as \( \mathcal{V}_{\text{train}}, \mathcal{V}_{\text{valid}}, \mathcal{V}_{\text{calib}}, \) and \( \mathcal{V}_{\text{test}} \). These methods presuppose that the distributions of calibration nodes \( \mathcal{V}_{\text{calib}} \) and test nodes \( \mathcal{V}_{\text{test}} \) are exchangeable during inference, meaning their joint distribution remains unchanged when samples are permuted. This assumption breaks down in federated graph settings for two primary reasons.

First, inherent dependencies among nodes due to their connectivities violate exchangeability if the test data is not present during training. Secondly, the distribution of graph data across different clients in a federated setting tends to vary, leading to non-exchangeable distributions. Specifically, the sets \( \mathcal{V}_{\text{calib}} \) and \( \mathcal{V}_{\text{test}} \) are not exchangeable, as their respective subsets \( \mathcal{V}_{\text{calib}}^{(k)} \) and \( \mathcal{V}_{\text{test}}^{(k')} \) may originate from distinct clients (\( k \neq k' \)). This variability underscores the challenges in assuming uniform data distribution across clients. For example, hospitals specializing in certain medical fields might predominantly treat patients from specific demographic groups, leading to skewed data distributions. Similarly, graph partitioning algorithms like METIS \citep{karypis1997metis}, used for simulating subgraph FL scenarios, aim to minimize edge cuts across partitions, often resulting in subgraphs that do not share the same data distribution.

\begin{table}[h]
\caption{Number of partitions (\( K \)) and missing edges (\( \Delta E \)).} 
\vspace{0.1cm}
\label{tab:missing-links}
\centering
\begin{tabular}{ccccc}
\toprule
Dataset & \( |E| \) & \( K \) & \( \Delta E \) & \( \Delta E \% \) \\
\midrule
\multirow{3}{*}{Cora} & \multirow{3}{*}{10,138} 
& 5 & 604 & 5.96\% \\
& & 10 & 806 & 7.95\% \\
& & 20 & 1,230 & 12.13\% \\
\midrule
\multirow{3}{*}{CiteSeer} & \multirow{3}{*}{7,358} 
& 5 & 310 & 4.21\% \\
& & 10 & 608 & 8.26\% \\
& & 20 & 848 & 11.52\% \\
\bottomrule
\end{tabular}
\end{table}

\textit{Missing Neighbor Information:} Another significant challenge in federated graph CP is the presence of missing neighbor information across client subgraphs. Consider a scenario where a patient visits multiple hospitals within the same city, maintaining separate records at each location. Due to conflicts of interest, it is impractical for hospitals to share their patient networks, leading to incomplete edge information in the overall graph. In simulations of federated learning based on graph partitioning, increasing the number of clients amplifies the number of missing links between them, as shown in Table~\ref{tab:missing-links}.

\begin{figure}[htbp]
    \centering
        \includegraphics[width=0.4\textwidth]{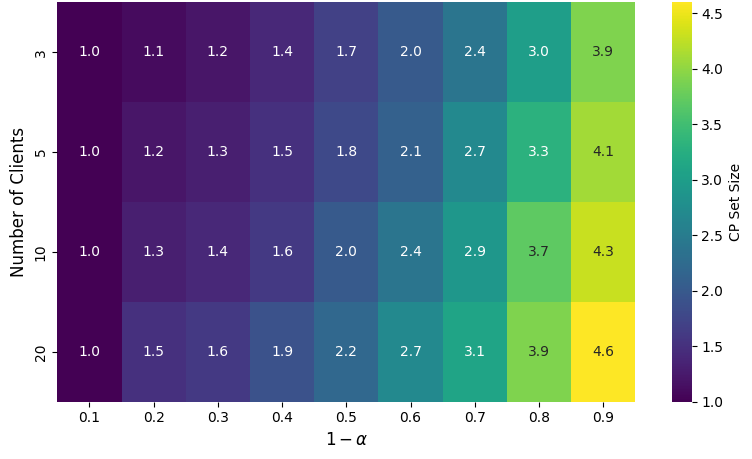}
    \caption{Effect of the number of clients on CP set size for the Cora dataset.}
    \label{fig:set-size}
\end{figure}

These missing edges, which carry critical neighborhood information, remain uncaptured by any single client subgraph. This absence becomes particularly problematic when CP techniques are applied to partitioned graph data, as it can impair model performance and increase the size of prediction sets due to insufficient coverage of the data's connectivity. Figure~\ref{fig:set-size} illustrates this issue, showing how the increasing number of missing links correlates with larger prediction set sizes through empirical evaluation.

Given these complexities, it is necessary to demonstrate how CP can be applied to non-exchangeable graph data and how the inefficiency caused by missing neighbor information can be mitigated within federated graph environments.

\section{Method} \label{sec:method}

\subsection{Partially Exchangeable Non-Conformity Scores}
\label{sec:method1}

To deploy Conformal Prediction for federated graph-structured data under a transductive learning setting, we need to ensure the exchangeability condition is met. We adopt the principle of partial exchangeability, as proposed by \citet{de1980condition} and applied to non-graph-based models by \citet{lu2023federated}. Specifically, we demonstrate that non-conformity scores within each client are permutation invariant when using a permutation-invariant GNN model for training under the transductive setting.

Consider a graph $\mathcal{G}^k = (\mathcal{V}^k, \mathcal{E}^k)$ at client $k$, where $\mathcal{V}^k$ denotes the set of nodes, $\mathcal{E}^k$ the set of edges, and each node $v \in \mathcal{V}^k$ has a feature vector $x_v \in \mathbb{R}^d$. The dataset includes distinct node subsets for training, validation, calibration, and testing: $\mathcal{V}^k_{\text{train}}$, $\mathcal{V}^k_{\text{valid}}$, $\mathcal{V}^k_{\text{calib}}$, and $\mathcal{V}^k_{\text{test}}$, respectively.

\begin{assumption}
\label{assumption:permutation-invariance}
Let $S$ be a global non-conformity score function learned in a federated setting, designed to be permutation invariant with respect to the calibration and test nodes within each client. For any permutation $\pi_k$ of client $k$'s calibration and test nodes, the non-conformity scores satisfy:
\[
\{ S(x_v, y_v) : v \in \mathcal{V}^k_{\text{calib}} \cup \mathcal{V}^k_{\text{test}} \} = 
\]
\[\{ S(x_{\pi_k(v)}, y_{\pi_k(v)}) : v \in \mathcal{V}^k_{\text{calib}} \cup \mathcal{V}^k_{\text{test}} \}.
\]
\end{assumption}

Non-conformity scores obtained through GNN training satisfy the above assumption because chosen GNN models are inherently permutation invariant with respect to node ordering. Each local GNN model accesses all node features during training and optimizes the objective function based solely on the training and validation nodes, which remain unchanged under permutation of the calibration and test nodes. Under Assumption~\ref{assumption:permutation-invariance}, we establish the following lemma.

\begin{lemma}
\label{lemma:invariance}
Within the transductive learning setting, assuming permutation invariance in graph learning over the unordered graph \(\mathcal{G}^k = (\mathcal{V}^k, \mathcal{E}^k)\), the set of non-conformity scores \(\{s_v\}_{v \in \mathcal{V}^k_{\text{calib}} \cup \mathcal{V}^k_{\text{test}}}\) is invariant under permutations of the calibration and test nodes.
\end{lemma}

The proof of Lemma~\ref{lemma:invariance} is provided in Appendix. Lemma~\ref{lemma:invariance} establishes the intra-client exchangeability of calibration and test samples for transductive node classification. Using Lemma~\ref{lemma:invariance}, we extend the concept of partial exchangeability to federated graph learning.

Assume that the subgraph at client $k$, $\mathcal{G}^k$, is sampled from a distribution $P_k$. During inference, a random test node $v_{\text{test}}$, with features and label $(x_{v_{\text{test}}}, y_{v_{\text{test}}})$, is assumed to be sampled from a global distribution $Q_{\text{test}}$, which is a mixture of the client subgraph distributions according to a probability vector $p$:
\[
Q_{\text{test}} = \sum_{k=1}^K p_k P_k,
\]
which essentially states that $v_{\text{test}}$ belongs to client $k$ with probability $p_k$.

\begin{definition}[Partial Exchangeability]
\label{def:partial-exchangeability}
Partial exchangeability in the context of federated learning assumes that the non-conformity scores between a test node and the calibration nodes within the same client are exchangeable, but this exchangeability does not necessarily extend to nodes from different clients.
\end{definition}

\begin{assumption}
\label{assumption:partial-exchangeability}
Consider a calibration set $\{v_i\}_{i=1}^{n_k}$ in client $k$ and a test node $v_{\text{test}}$ in the same client. Under the framework of partial exchangeability (Definition~\ref{def:partial-exchangeability}), the non-conformity scores \(s_{v_{\text{test}}}\) and \(\{s_{v_i}\}_{i=1}^{n_k}\) are assumed to be exchangeable with probability $p_k$, consistent with Assumption~\ref{assumption:permutation-invariance}. Therefore, $v_{\text{test}}$ is partially exchangeable with all calibration nodes within client $k$.
\end{assumption}

This assumption is justified by the properties of our non-conformity score function \(S\), which, as established under Assumption~\ref{assumption:permutation-invariance}, is designed to be permutation invariant within each client's data. This property supports the hypothesis that within a client, the test node and calibration nodes can be considered exchangeable in terms of their non-conformity scores. The limitation to within-client exchangeability is due to potential differences in data distribution across different clients, which Assumption~\ref{assumption:permutation-invariance} does not necessarily overcome. This limitation modifies the upper bound of the coverage guarantee, as elucidated in Theorem~\ref{theorem:coverage}. Details of this assumption can be found in Appendix.

\begin{theorem}
\label{theorem:coverage}
Suppose there are $n_k$ calibration nodes in client $k$'s subgraph. Let $N = \sum_{k=1}^K n_k$ and assume $p_k = (n_k + 1)/(N + K)$. If the non-conformity scores are arranged in non-decreasing order as $\{ S_{(1)}, S_{(2)}, \dots, S_{(N+K)} \}$, then the $\alpha$-quantile, $\hat{q}_{\alpha}$, is the $\lceil (1-\alpha)(N+K) \rceil$-th smallest value in this set. Consequently, the prediction set
\[
C_{\alpha}(v_{\text{test}}) = \{ y \in \mathcal{Y} \mid S(x_{v_{\text{test}}}, y) \leq \hat{q}_{\alpha} \}
\]
is a valid conformal predictor where:
\[
1 - \alpha \leq P\left( y_{\text{test}} \in C_{\alpha}(x_{v_{\text{test}}}) \right) \leq 1 - \alpha + \frac{K}{N+K}.
\]
\end{theorem}

This theorem ensures that our method achieves at least $(1 - \alpha)$ marginal coverage. The proof is provided in Appendix.

\subsection{Generating Representative Node Features with Variational Autoencoders}\label{sec:method2}

\begin{figure*}[htbp]
    \centering
        \includegraphics[width=0.78\linewidth]{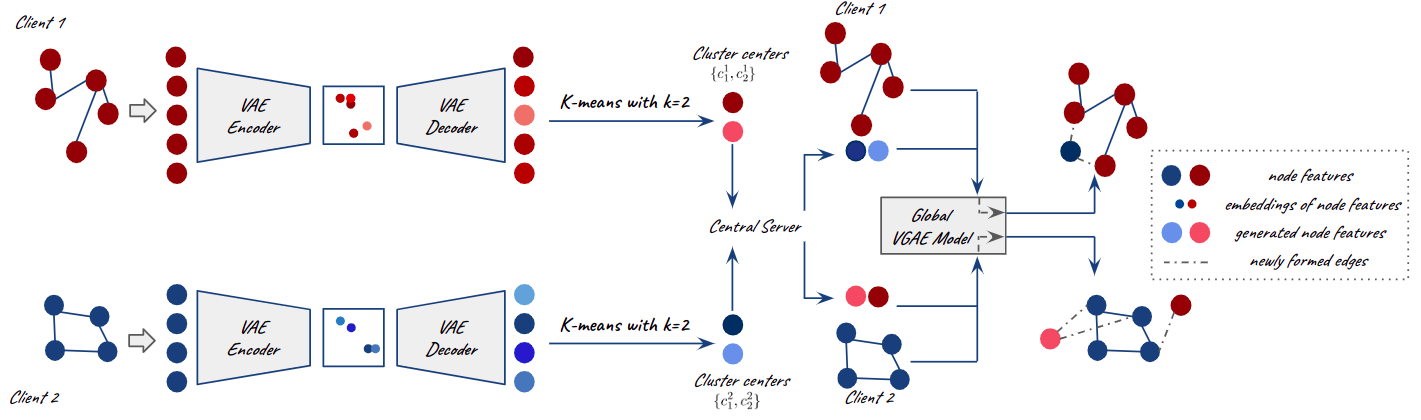}
    \caption{\textbf{Missing neighbor generation framework.} Clients use \texttt{VAE}s to generate node features, apply K-means clustering for prototype selection, and share these with a central server. The server redistributes prototypes to enrich client subgraphs, aiding in link prediction with a global \texttt{VGAE} model.}
 \label{fig-framework}
\end{figure*}

To mitigate the issue of missing neighbor information in federated graph learning, we introduce a novel approach that utilizes VAEs to generate representative node features within each client. These generated features are shared with the central server and then broadcast across clients to enrich the local subgraphs, thereby addressing the problem of missing links.

Each client \( k \) trains a \texttt{VAE} on its local node features \( \{ x_v \}_{v \in \mathcal{V}^k_{\text{train}}} \subset \mathbb{R}^d \), aiming to capture the underlying distribution \( P_k \) of its data. The \texttt{VAE} consists of an encoder \( q_{\phi_k}(z|x) \) and a decoder \( p_{\theta_k}(x|z) \), where \( z \in \mathbb{R}^{d'} \) is the latent representation, with \( d' < d \). The \texttt{VAE} is trained by maximizing the ELBO given in \ref{sec:vae}.

After training, each client generates reconstructed node features by passing its original node features through the encoder and decoder:

\[z_v = q_{\phi_k}(x_v), \quad \tilde{x}_v = p_{\theta_k}(z_v), \quad \forall v \in \mathcal{V}^k_{\text{train}}.
\]

Next, K-Means clustering \citep{kodinariya2013review} is applied to the reconstructed node features \( \{ \tilde{x}_v \} \) to identify \( M_k \) cluster centers \( \{ c_m^k \}_{m=1}^{M_k} \subset \mathbb{R}^d \):
\[
c_m^k = \frac{1}{|C_m^k|} \sum_{\tilde{x}_v \in C_m^k} \tilde{x}_v,
\]
where \( C_m^k \) is the set of reconstructed node features assigned to cluster \( m \) in client \( k \). The number of clusters \( M_k \) is determined experimentally through hyperparameter tuning.

The cluster centers \( \{ c_m^k \} \) are then used as prototype features and shared with the central server. The server aggregates the prototype features from all clients and broadcasts them back to each client. This process allows clients to augment their local subgraphs with representative node features from other clients, effectively approximating the missing neighbor information.

\subsection{Link Prediction with VGAE for Missing Neighbor Completion}
\label{sec:method3}

After the generated node features are collected by the central server and broadcast to the clients, we need to predict possible edge formations between the generated nodes and the client subgraphs. To this end, we employ a Variational Graph Autoencoder, effective in graph reconstruction tasks, suitable for our graph completion problem. The \texttt{VGAE} model is trained to minimize the ELBO loss.

To ensure that our link prediction model generalizes well across all client subgraphs, we train the \texttt{VGAE} in a federated setting using the \texttt{FedAvg} \citep{sun2022decentralized} algorithm. Different client subgraphs may have varying connectivity patterns; thus, the model needs to generalize to diverse subgraphs.

After training, the \texttt{VGAE} model is used for link prediction between generated nodes \(\hat{X}\) and local subgraph nodes \(X^k\). For each client \(k\), the link prediction process is as follows:

\begin{enumerate}
    \item \textbf{Compute edge probabilities} between generated nodes and local nodes:
    \(
    \hat{P}^k = \text{\texttt{VGAE}}(\hat{X}, X^k).
    \)
    \item \textbf{Select the top \(p\%\) of edge probabilities} to form new edges:
    \(
    \mathcal{E}^k \coloneqq \mathcal{E}^k \cup \left\{ (u,v) \mid (u,v) \in \text{Top}_p(\hat{P}^k) \right\}.
    \)
    \item \textbf{Update the node set and features}:
    \(
    \mathcal{V}^k \coloneqq \mathcal{V}^k \cup \hat{\mathcal{V}}, \quad X^k \coloneqq X^k \cup \hat{X}.
    \)
\end{enumerate}

Here, \(\text{Top}_p(\hat{P}^k)\) denotes the set of edges corresponding to the highest \(p\%\) of predicted edge probabilities in \(\hat{P}^k\). By integrating these new edges and nodes into their local subgraphs, clients enhance their models with previously missing neighbor information. This process is summarized in the Algorithm provided in Appendix.

\section{Experiments} \label{sec:experiments}

We conduct experiments on four real-world datasets to demonstrate the effectiveness of our proposed federated conformal prediction method on graph data with varying numbers of clients. 

\subsection{Experimental Setup}

We evaluate our method on four widely used graph datasets: Cora, CiteSeer, PubMed \citep{yang2016revisiting}, and Amazon Computers \citep{shchur2018pitfalls}. To simulate a federated learning environment, we partition each graph into clusters of $K = 3$, $5$, $10$, and $20$ using the METIS graph partitioning algorithm \citep{karypis1997metis}, which ensures clusters are of similar sizes and minimizes edge cuts between partitions. This partitioning introduces missing links between subgraphs, reflecting real-world scenarios where data is distributed across different clients with incomplete neighbor information.

We implement two-layer local permutation-invariant GNN models, GCNs or GraphSAGE with mean pooling, and employ the FedAvg algorithm \citep{mcmahan2017communication} to train the global GNN model. The batch size and learning rate for training local GNNs are set to 32 and 0.01, respectively, using the Adam optimizer. For the VAE and VGAE, we use the official implementations from the PyTorch Geometric package \citep{fey2019fast}, with hidden dimensions of size 64 and 16, respectively. The hyperparameters for percentages and number of clusters  were determined through a grid search over \( p \in \{0.01, 0.02, 0.04, 0.08, 0.10\} \) and \( M \in \{2, 5, 10, 20\} \). The VAE decoder mirrors the encoder dimensions, while the VGAE utilizes an inner product decoder.

Each local subgraph is divided into training, calibration, and test sets in a 20\%/40\%/40\% ratio. 20\% of the training set is used for validation. All experiments are conducted on NVIDIA RTX A5000-24GB GPUs.

As recommended by \citet{lu2023federated}, we apply temperature scaling to the conformal procedure. We average the locally learned temperatures across clients before initiating the federated conformal procedure.

To estimate the set-valued function \( C_\alpha \) within our framework, we compute the quantile of conformal scores \( \{s^k_i\}_{i=1}^{n_k} \) for each client \( k \in [K] \), where each score \( s^k_i = S(x^k_i, y^k_i) \) is distributed across \( K \) clients. We employ distributed quantile estimation techniques, which have proven effective in traditional federated conformal prediction settings \citep{lu2023federated}. Specifically, we adopt quantile averaging \citep{luo2016quantiles} and T-Digest \citep{dunning2021t}, a quantile sketching algorithm designed for efficient online quantile estimation in distributed settings.

We present our main results using the APS non-conformity score (with quantile averaging). We report the average conformal prediction set sizes across clients after local training of GNN models on each client (denoted as \texttt{Loc} in Table~\ref{results1}). This allows us to evaluate the empirical impact of the federated conformal procedure on graph data. The \texttt{Fed} entry in the table presents results from the experimental validation of our federated conformal prediction method. Results from our generative framework are indicated as \texttt{Gen}.

\begin{table*}[h]
\centering
\vspace{-0.15cm}
\caption{Conformal prediction (CP) set size comparison on four datasets with partition numbers \( K = 3, 5, 10, \) and \( 20 \) using the APS non-conformity score. Set sizes are presented for confidence levels \( 1 - \alpha = 0.95, 0.90, \) and \( 0.80 \). The corresponding standard deviations are given, averaged over 5 runs.}
\vspace{0.1cm}
\label{results1}
\setlength{\tabcolsep}{2pt}
\begin{tabular}{@{}lcccc cccc@{}}
\toprule
& \multicolumn{4}{c}{\textbf{Cora}} 
& \multicolumn{4}{c}{\textbf{PubMed}} \\
\cmidrule(lr){2-5} \cmidrule(lr){6-9}
\textbf{Model} & \textbf{\( K=3 \)} & \textbf{\( K=5 \)} & \textbf{\( K=10 \)} & \textbf{\( K=20 \)} & \textbf{\( K=3 \)} & \textbf{\( K=5 \)} & \textbf{\( K=10 \)} & \textbf{\( K=20 \)} \\
\midrule
\texttt{Loc (0.95)} & 4.97 \std{0.02} & 5.27 \std{0.02} & 5.53 \std{0.02} & 5.90 \std{0.02} 
        & 1.94 \std{0.03} & 1.97 \std{0.03} & 1.98 \std{0.04} & 2.04 \std{0.01}  \\
        
\texttt{Fed (0.95)}   & 4.31 \std{0.02} & \textbf{4.94} \std{0.02} & 5.02 \std{0.02} & 5.79 \std{0.02} 
        & 1.80 \std{0.02} & 1.78 \std{0.02} & 1.83 \std{0.01} & 1.77 \std{0.02}  \\
        
\texttt{Gen (0.95)}   & \textbf{4.25} \std{0.02} & 5.09 \std{0.02} & \textbf{4.86} \std{0.02} & \textbf{5.40} \std{0.02} 
        & \textbf{1.72} \std{0.02} & \textbf{1.78} \std{0.01} & \textbf{1.80} \std{0.02} & \textbf{1.69} \std{0.02}  \\
        
               \cmidrule(lr){2-5} \cmidrule(lr){6-9}
               
\texttt{Loc (0.90)} & 4.12 \std{0.01} & 4.54 \std{0.01} & 4.83 \std{0.03} & 4.99 \std{0.03} 
        & 1.79 \std{0.00} & 1.86 \std{0.03} & 1.90 \std{0.03} & 1.95 \std{0.03}  \\
        
\texttt{Fed (0.90)}   & 3.34 \std{0.03} & 4.14 \std{0.03} & 4.32 \std{0.02} & 4.13 \std{0.01}
        & 1.61 \std{0.03} & \textbf{1.60} \std{0.01} & 1.68 \std{0.01} & 1.53 \std{0.02}  \\
        
\texttt{Gen (0.90)}   & \textbf{3.34} \std{0.02} & \textbf{4.10} \std{0.02} & \textbf{3.98} \std{0.01} & \textbf{3.90} \std{0.04} 
        & \textbf{1.55} \std{0.01} & \textbf{1.60} \std{0.01} & \textbf{1.62} \std{0.02} & \textbf{1.49} \std{0.01}  \\
        
               \cmidrule(lr){2-5} \cmidrule(lr){6-9}
               
\texttt{Loc (0.80)} & 3.17 \std{0.01} & 3.77 \std{0.01} & 3.87 \std{0.02} & 4.14 \std{0.02} 
        & 1.72 \std{0.01} & 1.78 \std{0.01} & 1.80 \std{0.01} & 1.69 \std{0.01}  \\
        
\texttt{Fed (0.80)}   & \textbf{2.45} \std{0.01} & \textbf{2.95 }\std{0.01} & 2.93 \std{0.03} & 3.17 \std{0.03} 
        & 1.55 \std{0.01} & 1.60 \std{0.02} & 1.62 \std{0.00} & 1.49 \std{0.02}  \\
        
\texttt{Gen (0.80)}   & 2.51 \std{0.03} & 2.98 \std{0.05} & \textbf{2.92} \std{0.02} & \textbf{2.88} \std{0.03} 
        & \textbf{1.41} \std{0.04} & \textbf{1.42} \std{0.01} & \textbf{1.45} \std{0.03} & \textbf{1.37} \std{0.03}  \\

\midrule
& \multicolumn{4}{c}{\textbf{CiteSeer}} 
& \multicolumn{4}{c}{\textbf{Computers}} \\
\cmidrule(lr){2-5} \cmidrule(lr){6-9}
\textbf{Model} & \textbf{\( K=3 \)} & \textbf{\( K=5 \)} & \textbf{\( K=10 \)} & \textbf{\( K=20 \)} & \textbf{\( K=3 \)} & \textbf{\( K=5 \)} & \textbf{\( K=10 \)} & \textbf{\( K=20 \)} \\
\midrule
\texttt{Loc (0.95)} & 4.80 \std{0.02} & 4.95 \std{0.02} & 4.99 \std{0.02} & 4.99 \std{0.03}
        & 6.18 \std{0.03} & 6.31 \std{0.04} & 6.71 \std{0.02} & 6.45 \std{0.02}  \\
        
\texttt{Fed (0.95)}   & 3.89 \std{0.04} & 4.12 \std{0.01} & \textbf{4.19} \std{0.04} & 4.42 \std{0.01} 
        & 5.58 \std{0.03} & 5.96 \std{0.02} & 4.76 \std{0.03} & 6.10 \std{0.02}  \\
        
\texttt{Gen (0.95)}   & \textbf{3.72} \std{0.02} & \textbf{4.11} \std{0.01} & 4.55 \std{0.02} & \textbf{4.27} \std{0.01} 
        & \textbf{5.08} \std{0.02} & \textbf{5.86} \std{0.03} & \textbf{4.21} \std{0.02} & \textbf{5.30} \std{0.01}  \\
        
               \cmidrule(lr){2-5} \cmidrule(lr){6-9}
               
\texttt{Loc (0.90)} & 4.14 \std{0.01} & 4.62 \std{0.01} & 4.73 \std{0.03} & 4.87 \std{0.03} 
        & 5.27 \std{0.04} & 5.26 \std{0.04} & 5.66 \std{0.02} & 5.73 \std{0.03}  \\
        
\texttt{Fed (0.90)}   & 3.16 \std{0.01} & 3.09 \std{0.02} & 3.14 \std{0.03} & 3.82 \std{0.02} 
        & 4.78 \std{0.04} & 5.37 \std{0.02} & 4.09 \std{0.02} & 5.46 \std{0.01}  \\
        
\texttt{Gen (0.90)}   & \textbf{2.95} \std{0.02} & \textbf{2.96} \std{0.01} & \textbf{3.08} \std{0.02} & \textbf{3.65} \std{0.01} 
        & \textbf{4.59} \std{0.03} & \textbf{5.09} \std{0.02} & \textbf{3.57} \std{0.01} & \textbf{4.69} \std{0.01}  \\
        
               \cmidrule(lr){2-5} \cmidrule(lr){6-9}
               
\texttt{Loc (0.80)} & 3.35 \std{0.01} & 3.59 \std{0.02} & 3.84 \std{0.03} & 3.98 \std{0.02} 
        & 4.24 \std{0.04} & 4.28 \std{0.01} & 4.55 \std{0.03} & \textbf{3.96 }\std{0.01}  \\
        
\texttt{Fed (0.80)}   & 2.45 \std{0.03} & 2.22 \std{0.02} & 2.17 \std{0.02} & 2.85 \std{0.01} 
        & \textbf{3.60} \std{0.01} & 4.28 \std{0.03} & 3.39 \std{0.01} & 4.66 \std{0.03}  \\
        
\texttt{Gen (0.80)}   & \textbf{2.19} \std{0.02} & \textbf{2.11} \std{0.02} & \textbf{2.14} \std{0.03} & \textbf{2.66} \std{0.02} 
        & 3.62 \std{0.03} & \textbf{3.97} \std{0.02} & \textbf{2.96} \std{0.02} & 3.97 \std{0.01}  \\

\bottomrule
\end{tabular}
\end{table*}

\subsection{Main Results and Efficiency Analysis}

Our experimental results, presented in Table~\ref{results1}, consistently demonstrate that federated training (\texttt{Fed}) achieves smaller conformal prediction set sizes compared to local training (\texttt{Loc}). This improvement is particularly pronounced as the number of clients increases. For example, on the PubMed dataset with \( K = 20 \) clients at a confidence level of \( 1 - \alpha = 0.95 \), the federated approach yields an average CP set size of \( 1.77 \pm 0.02 \), compared to \( 2.04 \pm 0.01 \) for local training, indicating a significant reduction in uncertainty.

Our missing neighbor generator (\texttt{Gen}) further refines the CP set sizes across varying configurations and datasets. This approach effectively mitigates the impact of incomplete neighbor data, allowing the conformal predictor to perform closer to its optimal state, where all connections are known. Notably, in the Computers dataset with \( K = 20 \) and a confidence level of \( 1 - \alpha = 0.95 \), the generative method achieves a CP set size of \( 5.30 \pm 0.01 \), improving upon both federated (\( 6.10 \pm 0.02 \)) and local training (\( 6.45 \pm 0.02 \)). As the number of clients increases, our method continues to return less uncertain sets compared to \texttt{Loc} and \texttt{Fed}, demonstrating the scalability of our approach.

\begin{figure}[htbp]
\vspace{0.2cm}
    \centering
        \includegraphics[width=0.46\textwidth]{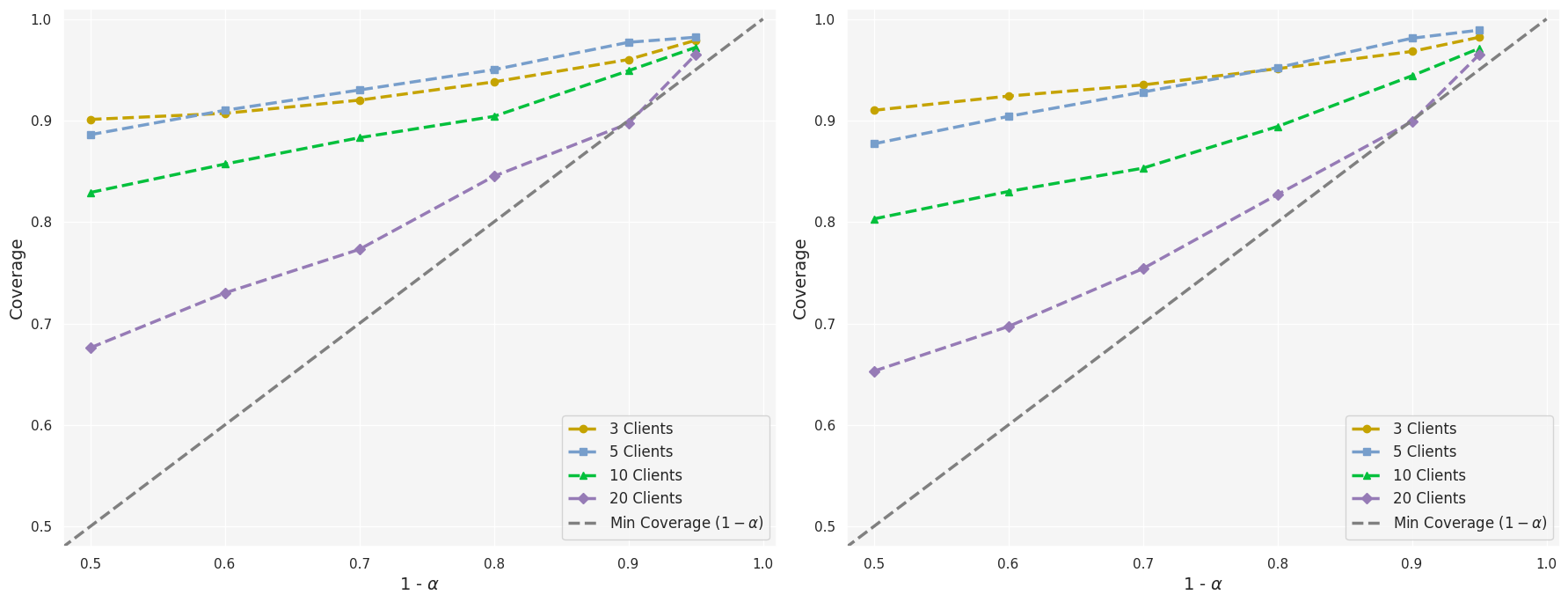}
        \caption{\textbf{Coverage Rates:} Coverage rates for \texttt{Fed} (left) and \texttt{Gen} (right) models across varying \( K \) on the Cora dataset.}
        \label{fig:coverage_rates}
\vspace{0.2cm}
\end{figure}

\subsection{Coverage Rates}

As shown in our theoretical analysis in Section~\ref{sec:method}, applying conformal prediction in federated graph learning maintains the lower bound for CP coverage. We experimentally verify this on the Cora dataset, as shown in Figure~\ref{fig:coverage_rates}. The figure displays the coverage rates for different numbers of clients, confirming that our method ensures the desired coverage rate. For lower \( 1 - \alpha \) values, the model easily achieves coverage rates above the threshold, and its performance is maintained even at more challenging miscoverage rates. This demonstrates that CP is a useful method for uncertainty quantification in high-stakes scenarios. Additionally, applying our generative model achieves the desired coverage rate while returning smaller sets.

\subsection{Case Study: Impact on Accuracy}

While our main focus is on CP efficiency, we also examine how our generative method affects the predictive accuracy. We compare our approach with \citet{zhang2021subgraph}, which directly aims at improving accuracy by completing missing subgraphs through \texttt{FedSage+}. We evaluate the improvement in terms of percentage change for both methods, as illustrated in Table~\ref{tab:accuracy}. Our method consistently improves the test accuracy of subgraph federated learning using GraphSAGE, and even outperforms \texttt{FedSage+} in some cases.

\begin{table}[t]
\small
\caption{Percentage change in accuracy for our method and \texttt{FedSage+} on Cora and CiteSeer datasets with varying \( K \).}\label{tab:accuracy}
\vspace{-0.2cm}
\centering
\resizebox{0.4\textwidth}{!}{
\begin{tabular}{l|cccc}
    \toprule
    {} & \( K=3 \) & \( K=5 \) & \( K=10 \) & \( K=20 \) \\
    \midrule
    \multicolumn{5}{c}{\textbf{Cora}} \\
    \cmidrule(lr){1-5}
    \texttt{FedSage+}
    & \color{MediumGreen} \textbf{+0.34\%} 
    & \color{MediumGreen} \textbf{+0.04\%} 
    & \color{MediumGreen} \textbf{+0.07\%} 
    &  \textbf{N/A} \\
    \texttt{Ours}
    & \color{DarkGreen} \textbf{+1.55\%} 
    & \color{MediumRed} \textbf{-0.91\%}
    & \color{DarkGreen} \textbf{+1.19\%} 
    & \color{DarkGreen} \textbf{+5.94\%} \\
    \midrule
    \multicolumn{5}{c}{\textbf{CiteSeer}} \\
    \cmidrule(lr){1-5}
    \texttt{FedSage+}
    & \color{DarkGreen} \textbf{+2.94\%} 
    & \color{DarkGreen} \textbf{+2.96\%} 
    & \color{DarkGreen} \textbf{+3.26\%}
    & \textbf{N/A} \\
    \texttt{Ours}
    & \color{MediumGreen} \textbf{+2.75\%} 
    & \color{MediumGreen} \textbf{+2.55\%}
    & \color{MediumGreen} \textbf{+1.13\%}
    & \color{DarkGreen} \textbf{+3.51\%} \\
    \bottomrule
\end{tabular}
}
\end{table}

\section{Conclusion} \label{sec:conclusion}

We introduced federated conformal prediction for node classification under transductive settings on graphs, establishing theoretical foundations and confirming reliable marginal coverage guarantees. Additionally, we introduced a generative model to address inefficiencies in CP set sizes due to missing links across client subgraphs. Experiments on real-world graphs demonstrate the practical effectiveness of our approach.

\newpage
\onecolumn
\aistatstitle{Supplementary Materials}

\section{Partial Exchangeability Proofs}
\label{appendix:1}

\subsection{Proof of Lemma 1}

Consider the unordered graph \(\mathcal{G}^k = (\mathcal{V}^k, \mathcal{E}^k)\) within the permutation-invariant graph learning environment as outlined in the conditions of Lemma 1. Assuming that the graph structure, attribute information, and node label information are fixed, we define the nonconformity scores at nodes in \(\mathcal{V}^k_{\text{calib}} \cup \mathcal{V}^k_{\text{test}}\) as
\[
\{s_v\} = S\left(\mathcal{V}^k, \mathcal{E}^k, \{(x_v, y_v)\}_{v \in \mathcal{V}^k_{\text{train}} \cup \mathcal{V}^k_{\text{valid}}}, \{x_v\}_{v \in \mathcal{V}^k_{\text{calib}} \cup \mathcal{V}^k_{\text{test}}}\right),
\]
where \(S\) denotes the scoring function used to compute the nonconformity scores.

Due to the permutation invariance of the model (Assumption 1), for any permutation \(\pi\) of the nodes in \(\mathcal{V}^k_{\text{calib}} \cup \mathcal{V}^k_{\text{test}}\), the nonconformity scores remain unchanged. Specifically, we have
\[
\{s_v\} = S\left(\pi\left(\mathcal{V}^k\right), \pi\left(\mathcal{E}^k\right), \{(x_v, y_v)\}_{v \in \mathcal{V}^k_{\text{train}} \cup \mathcal{V}^k_{\text{valid}}}, \{x_{\pi(v)}\}_{v \in \mathcal{V}^k_{\text{calib}} \cup \mathcal{V}^k_{\text{test}}}\right).
\]
Here, \(\pi\left(\mathcal{V}^k\right)\) and \(\pi\left(\mathcal{E}^k\right)\) denote the vertex set and edge set permuted according to \(\pi\).

This invariance implies that, regardless of the permutation of nodes in \(\mathcal{V}^k_{\text{calib}} \cup \mathcal{V}^k_{\text{test}}\), the computed nonconformity scores \(\{s_v\}\) remain the same. Therefore, the unordered set of scores \(\{s_v\}_{v \in \mathcal{V}^k_{\text{calib}} \cup \mathcal{V}^k_{\text{test}}}\) is invariant under permutations of the nodes, confirming the lemma's assertion about the stability and invariance of the score set in this setting.

\subsection{Remark on Assumption 2}

Under Assumption 2, the nonconformity scores \(\{s_{v_i}\}_{v_i \in \mathcal{V}^k_{\text{calib}}}\) for client \(k\) are identically distributed and exchangeable. Extending this set to include the score \(s_{v_{\text{test}}} = S(x_{v_{\text{test}}}, y_{v_{\text{test}}})\), where \((x_{v_{\text{test}}}, y_{v_{\text{test}}}) \sim P_k\) (the distribution for client \(k\)), the augmented set \(\{s_{v_i}\}_{v_i \in \mathcal{V}^k_{\text{calib}}} \cup \{s_{v_{\text{test}}}\}\) remains identically distributed and exchangeable.

This demonstrates that \(s_{v_{\text{test}}}\) is equivalent in distribution to any \(s_{v_i}\) in the calibration set. Therefore, the test score \(s_{v_{\text{test}}}\) can be considered as an additional sample from the same distribution, affirming the IID and exchangeability conditions outlined in Assumption 2.

\subsection{Proof of Theorem 1}

We aim to show that under the given assumptions, the conformal prediction framework achieves the intended coverage guarantees.

Let \(N = \sum_{k=1}^K n_k\) be the total number of calibration nodes across all clients, where \(n_k\) is the number of calibration nodes for client \(k\). Define \(p_k = \dfrac{n_k + 1}{N + K}\), so that \(\sum_{k=1}^K p_k = 1\).

For each client \(k\), let \(m_k(q)\) denote the number of nonconformity scores less than or equal to \(q\) among the \(n_k + 1\) scores (including the test node), that is,
\[
m_k(q) = \left|\left\{s_v \mid s_v \leq q, \, v \in \mathcal{V}^k_{\text{calib}} \cup \{v_{\text{test}}\}\right\}\right|.
\]

\newpage

Recall that the conformal quantile \(\hat{q}_\alpha\) is defined as the \(\lceil (1 - \alpha)(N + K) \rceil\)-th smallest nonconformity score among all calibration scores and test scores from all clients. Thus,
\[
\sum_{k=1}^K m_k(\hat{q}_\alpha) = \lceil (1 - \alpha)(N + K) \rceil.
\]

Define the event \(\mathcal{E}\) as the combined ordering of nonconformity scores within each client, that is,
\[
\mathcal{E} = \left\{ \forall k \in [K], \text{ the nonconformity scores } \{s^k_i\}_{i=1}^{n_k + 1} \text{ are in a fixed order} \right\},
\]
where \(\{s^k_i\}_{i=1}^{n_k + 1}\) are the nonconformity scores for client \(k\), including the test score, sorted in some fixed order.

Conditioned on \(\mathcal{E}\), the number of scores less than or equal to \(\hat{q}_\alpha\), \(m_k(\hat{q}_\alpha)\), is deterministic for each client \(k\).

Under the exchangeability assumption, the probability that the test score \(s_{v_{\text{test}}}\) is less than or equal to \(\hat{q}_\alpha\) conditioned on \(\mathcal{E}\) is
\[
P(s_{v_{\text{test}}} \leq \hat{q}_\alpha \mid \mathcal{E}) = \sum_{k=1}^K p_k \cdot \frac{m_k(\hat{q}_\alpha)}{n_k + 1}.
\]

Therefore, we have
\[
P(s_{v_{\text{test}}} \leq \hat{q}_\alpha \mid \mathcal{E}) = \frac{\sum_{k=1}^K m_k(\hat{q}_\alpha)}{N + K} = \frac{\lceil (1 - \alpha)(N + K) \rceil}{N + K} \geq 1 - \alpha.
\]

Similarly, we can derive an upper bound:
\[
P(s_{v_{\text{test}}} \leq \hat{q}_\alpha \mid \mathcal{E}) \leq \frac{\sum_{k=1}^K (m_k(\hat{q}_\alpha) + 1)}{N + K} = \frac{\lceil (1 - \alpha)(N + K) \rceil + K}{N + K} \leq 1 - \alpha + \frac{K}{N + K}.
\]

Thus, we have established that the coverage probability satisfies
\[
1 - \alpha \leq P(s_{v_{\text{test}}} \leq \hat{q}_\alpha \mid \mathcal{E}) \leq 1 - \alpha + \frac{K}{N + K}.
\]

Since \(\mathcal{E}\) has probability 1 (it conditions on the ordering which is always possible), the unconditional probability satisfies the same bounds. This completes the proof that the conformal predictor maintains the desired coverage level under the partial exchangeability and permutation invariance assumptions in the graph-structured federated learning setting.

\section{Model Details and Detailed Algorithm}

The subsequent sections detail the algorithms employed in our proposed methodology, encompassing node generation, edge formation, and the application of CP to federated node classification tasks. 

\begin{algorithm}[H]
\caption{Federated Graph Learning with Missing Neighbor Generation and Conformal Prediction}
\label{alg:missing-neighbor-generation-conformal-prediction}
\begin{algorithmic}[1]
\Require 
    \( K \): Number of clients \\
    \( \{ (\mathcal{V}^k_{\text{train}}, X^k_{\text{train}}, \mathcal{E}^k) \}_{k=1}^K \): Local datasets \\
    \( M_k \): Number of clusters per client \\
    \( p\% \): Top percentage for edge selection \\
    \( R \): Number of federated rounds \\
    Learning rates, other hyperparameters

\Ensure 
    Augmented local graphs \( \{ \mathcal{G}^k = (\mathcal{V}^k, X^k, \mathcal{E}^k) \}_{k=1}^K \)
\Statex

\State \textbf{Step 1: Generate prototype node features}
\For{each client \( k = 1 \) to \( K \) \textbf{in parallel}}
    \State Train VAE \( q_{\phi_k}(z|x) \), \( p_{\theta_k}(x|z) \)
    \State Reconstruct features \( \tilde{x}_v = p_{\theta_k}(q_{\phi_k}(x_v)) \)
    \State Cluster \( \{ \tilde{x}_v \} \) into \( M_k \) centers \( \{ c_m^k \} \)
    \State Send \( \{ c_m^k \} \) to the server
\EndFor
\Statex
\State \textbf{Step 2: Aggregate and broadcast prototypes}
\State Aggregate \( \hat{X} = \bigcup_{k=1}^K \{ c_m^k \} \)
\State Broadcast \( \hat{X} \) to all clients
\Statex
\State \textbf{Step 3: Federated training of VGAE}
\State Initialize global VGAE parameters \( \Theta \)
\For{each round \( r = 1 \) to \( R \)}
    \For{each client \( k = 1 \) to \( K \) \textbf{in parallel}}
        \State Receive \( \Theta \)
        \State Augment \( X^k \leftarrow X^k_{\text{train}} \cup \hat{X} \)
        \State Train local VGAE \( q_{\psi_k}(Z|X^k, \mathcal{E}^k) \), \( p_{\varphi_k}(\mathcal{E}^k|Z) \)
        \State Send updated \( \Theta_k \) to server
    \EndFor
    \State Aggregate \( \Theta \leftarrow \frac{1}{K} \sum_{k=1}^K \Theta_k \)
\EndFor
\Statex
\State \textbf{Step 4: Link prediction and graph update}
\For{each client \( k = 1 \) to \( K \)}
    \State Compute edge probabilities \( \hat{P}^k = \text{VGAE}_\Theta(X^k, \mathcal{E}^k) \)
    \State Select top \( p\% \) edges to form new set \( \hat{\mathcal{E}}^k \)
    \State Update \( \mathcal{E}^k \leftarrow \mathcal{E}^k \cup \hat{\mathcal{E}}^k \)
\EndFor
\Statex
\State \textbf{Step 5: Final graph for downstream tasks}
\For{each client \( k = 1 \) to \( K \)}
    \State \( \mathcal{G}^k = (\mathcal{V}^k, X^k, \mathcal{E}^k) \)
    \State Proceed with downstream tasks
\EndFor
\Statex
\State \textbf{Step 6: Federated Conformal Prediction (Optional)}
\For{each client \( k = 1 \) to \( K \)}
    \State Compute local validation scores \( S^k_{\text{val}} \) and targets \( Y^k_{\text{val}} \)
    \State Tune temperature \( T \) based on validation data
    \State Compute conformal scores \( q_{\text{LAC}}^k, q_{\text{APS}}^k, q_{\text{RAPS}}^k \) using \( S^k_{\text{val}} \), \( Y^k_{\text{val}} \), and significance level \( \alpha \)
    \State Share local conformal scores with the server
\EndFor
\State Aggregate global conformal scores \( q_{\text{LAC}}, q_{\text{APS}}, q_{\text{RAPS}} \) on the server
\State Apply conformal prediction sets using \( q_{\text{LAC}}, q_{\text{APS}}, q_{\text{RAPS}} \) for test data
\end{algorithmic}
\end{algorithm}

\newpage

\subsection{Sparsity Regularization for Node Feature Generation}

In addition to the standard reconstruction and KL-divergence losses in the \texttt{VAE}, we incorporate a sparsity regularization term to encourage the generated node features to reflect the sparse nature of real-world graph data. This is crucial for datasets where most node features are inherently sparse, ensuring that the latent representations and reconstructed features remain close to the original sparse structure.

Given the latent representations \( z \in \mathbb{R}^{d'} \), the sparsity regularization is applied to the encoder activations to control the average activation levels across the latent dimensions. Let \( \hat{\rho} \in \mathbb{R}^{d'} \) denote the mean activation of the latent variables \( z \) over all nodes, defined as:

\[
\hat{\rho}_i = \frac{1}{|\mathcal{V}|} \sum_{v \in \mathcal{V}} z_{v,i}, \quad \forall i \in [1, d'].
\]

We introduce a sparsity target \( \rho \in (0,1) \) that specifies the desired level of activation for each latent variable. The sparsity loss \( \mathcal{L}_{\text{sparse}} \) is then defined as the Kullback-Leibler divergence between the desired activation \( \rho \) and the average activation \( \hat{\rho} \):

\[
\mathcal{L}_{\text{sparse}} = \sum_{i=1}^{d'} \left( \rho \log \frac{\rho}{\hat{\rho}_i} + (1 - \rho) \log \frac{1 - \rho}{1 - \hat{\rho}_i} \right).
\]

This loss term encourages the activations to stay close to the sparsity target \( \rho \), penalizing deviations from this target. A scaling factor \( \beta \) is used to adjust the contribution of this term, and the overall loss function for training the \texttt{VAE} becomes:

\[
\mathcal{L} = \lambda_{\text{rec}} \mathcal{L}_{\text{rec}} + \lambda_{\text{kl}} \mathcal{L}_{\text{kl}} + \beta \mathcal{L}_{\text{sparse}},
\]
where \( \mathcal{L}_{\text{rec}} \) is the reconstruction loss, \( \mathcal{L}_{\text{kl}} \) is the KL-divergence loss, and \( \lambda_{\text{rec}} \), \( \lambda_{\text{kl}} \), and \( \beta \) are weights controlling the relative importance of each term.

Incorporating this sparsity regularization helps ensure that the generated node features remain representative of the original sparse input data, improving the quality and fidelity of the reconstructed features in graph-based learning tasks.

\section{Complexity Analysis}
\label{sec:complexity}

In this section, we provide a complexity analysis of the proposed method, focusing on the communication overhead between the clients and the central server, as well as the computational cost related to the exchange and utilization of generated node features.

\subsection{Prototype Sharing Complexity}
After training the \texttt{VAE}, each client \( k \) identifies \( M_k \) cluster centers, representing the prototype features that will be shared with the central server. The dimensionality of each prototype is \( d \), and the total communication cost of sending the prototype features from client \( k \) to the server is:

\[
\mathcal{O}(M_k \cdot d).
\]

Since there are \( K \) clients in total, the overall communication complexity for sending prototypes to the server is:

\[
\mathcal{O}(K \cdot M_k \cdot d),
\]

where \( M_k \) may vary across clients but is typically constant for simplicity.

\subsection{Server Aggregation Complexity}
The central server aggregates the prototype features from all clients, combining them into a global set of features \( \hat{X} = \bigcup_{k=1}^K \{c_m^k\} \). This aggregation step involves concatenating the received prototypes, which has a complexity of:

\[
\mathcal{O}(K \cdot M_k \cdot d).
\]

The server then broadcasts the aggregated prototypes back to all clients. The communication complexity of broadcasting the prototypes from the server to all clients is:

\[
\mathcal{O}(K \cdot M_k \cdot d),
\]

assuming all clients receive the same set of \( (K-1) \cdot M_k \) prototypes. Thus, the total communication cost for the prototype-sharing phase (sending prototypes to the server and broadcasting them back) is:

\[
\mathcal{O}(2 \cdot K \cdot M_k \cdot d).
\]

\subsection{Federated Training Communication Complexity}
During the federated training of the \texttt{VGAE} model, each client \( k \) sends its local model updates \( \Theta_k \) to the central server. The model parameters \( \Theta_k \) are of size \( |\Theta| \), which is the same across all clients. The communication complexity for each client sending its updated model to the server is:

\[
\mathcal{O}(|\Theta|).
\]

The server aggregates the model updates from all \( K \) clients, which involves summing the model parameters. The complexity of this aggregation step is:

\[
\mathcal{O}(K \cdot |\Theta|).
\]

The server then sends the updated global model back to each client, with a communication complexity of:

\[
\mathcal{O}(K \cdot |\Theta|),
\]

since each client receives the full set of model parameters. Thus, the total communication complexity for one round of federated training is:

\[
\mathcal{O}(2 \cdot K \cdot |\Theta|).
\]

\subsection{Overall Communication Complexity}
The overall communication complexity of the proposed method consists of two main components: (1) prototype sharing and (2) federated training. The total communication complexity is the sum of these two components, which can be expressed as:

\[
\mathcal{O}(2 \cdot K \cdot M_k \cdot d + 2 \cdot K \cdot |\Theta| \cdot R).
\]

This complexity scales linearly with the number of clients \( K \), the number of prototypes \( M_k \), the number of training epochs \(R\), and the size of the model \( |\Theta| \). Therefore, the communication overhead remains manageable, even as the number of clients and the model size increase.

\section{Differential Privacy Analysis}\label{sec:dp}

In this section, we explore the integration of differential privacy (DP) into the node generation process of our federated learning setup. To ensure the privacy of individual node features, we train the \texttt{VAE}-based node generator using \(\epsilon\)-\(\delta\) differential privacy \citep{dwork2014algorithmic}. Under DP, a randomized mechanism \(\mathcal{M}\) satisfies \(\epsilon\)-\(\delta\) privacy if for any two neighboring datasets \(D\) and \(D'\), the following holds:

\[
\Pr[\mathcal{M}(D) \in S] \leq e^\epsilon \Pr[\mathcal{M}(D') \in S] + \delta,
\]
where \(\epsilon > 0\) controls the privacy loss and \(\delta\) accounts for the probability of a privacy breach.

The node generator model is trained using the Opacus library to implement privacy-preserving stochastic gradient descent (DP-SGD), which ensures that each client’s data is protected by clipping gradients and adding Gaussian noise. This technique introduces an additional noise term to the training process, making it difficult for an adversary to infer individual node features while still enabling useful feature generation. The impact of DP on model performance is explored by varying the privacy budget \(\epsilon\) and fixing \(\delta = 10^{-5}\).

\subsection{Performance with Varying Privacy Budgets}

We evaluate the effectiveness of our node generation method under different privacy budgets by training the \texttt{Gen} method with \(\epsilon\) values ranging from 1 to 25. We analyze the impact of privacy noise on the RAPS non-conformity scores for various \(1-\alpha\) values (ranging from 0.5 to 0.95), comparing the results against the non-private \texttt{Fed} and \texttt{Gen} methods.

\begin{figure}[htbp]
    \centering
    \includegraphics[width=0.7\linewidth]{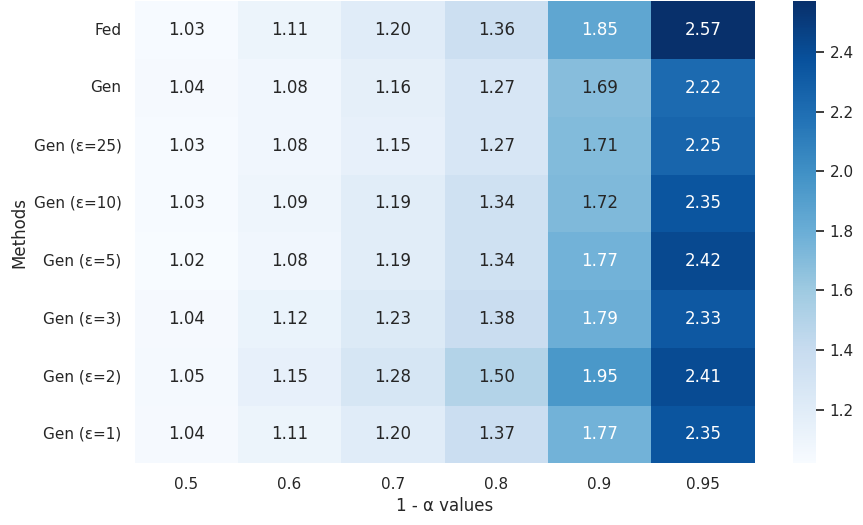}
    \caption{\textbf{Heatmap showing RAPS non-conformity scores for \texttt{Fed} and \texttt{Gen} methods across various \(\epsilon\)-values and \(1-\alpha\) values on 3 client Cora dataset.}}
    \label{fig:raps_heatmap}
\end{figure}

Figure \ref{fig:raps_heatmap} presents the observed scores. We note that as the privacy budget decreases (i.e., smaller \(\epsilon\) values), the performance of the \texttt{Gen} method degrades slightly, particularly for larger \(1-\alpha\) values. This degradation is expected due to the additional noise introduced by the DP mechanism, which affects the accuracy of the generated node features. However, even with \(\epsilon = 1\), the degradation remains relatively small, demonstrating that our approach maintains robust performance under strict privacy constraints.

From the results, we observe that at \(\epsilon=10\), the privacy-preserving \texttt{Gen} model closely approximates the performance of the non-private \texttt{Gen} method across all \(1-\alpha\) values. However, with stricter privacy budgets (e.g., \(\epsilon=1\)), there is a marginal increase in non-conformity scores, indicating a slight decrease in accuracy due to the added noise. Despite this, the model remains competitive even under the strictest privacy constraints.

\subsection{Conclusion}

Our experiments show that incorporating \(\epsilon\)-\(\delta\) differential privacy into the node generation process enables strong privacy guarantees with minimal impact on performance. Even under the strictest privacy settings, the model retains its ability to generate useful node features, as evidenced by the modest increases in RAPS non-conformity scores.

\section{Datasets Statistics}
\label{appendix:3}

We used the largest connected components of Cora, CiteSeer, PubMed \citep{yang2016revisiting}, and Amazon Computers \citep{shchur2018pitfalls} datasets in the Pytorch Geometric package \citep{fey2019fast}. Dataset statistics are given in Table \ref{datasets}.

\begin{table}[h]
\centering
\caption{Dataset statistics.} 
\label{datasets}
\begin{tabular}{lcccc}
\hline
\textbf{Dataset}   & \textbf{\# Nodes} & \textbf{\# Edges} & \textbf{\# Features} & \textbf{\# Labels} \\ \hline
Cora               & 2485             & 10138            & 2485                & 7                  \\
CiteSeer           & 2120             & 7358            & 3703                  & 6                  \\
PubMed               & 19717            & 88648           & 500                & 3                  \\
Computers                 & 13752           & 491722           & 767                & 10                 \\ \hline
\end{tabular}
\end{table}

\section{Comparison of Non-Conformity Scores}
\label{appendix:5}

Regularized Adaptive Prediction Sets (RAPS) \citep{angelopoulos2020uncertainty} refine APS by introducing regularization to penalize less likely labels. RAPS modifies the score function to include a regularization term, encouraging smaller prediction sets. The score function is defined as 
\[s(x, y) = - (\rho(x, y) + u \cdot \pi(x)y + \nu \max(o(x, y) - k, 0))
\], where $\nu$ and $k$ are hyperparameters, and $o(x, y)$ represents the rank of $y$.

Least Ambiguous Set-Valued Classifiers (LAC) \citep{sadinle2019least}assess classification uncertainty. The classifier's score, \( s(x, y) \), is given by:
\[
s(x, y) = 1 - [f(x)]_y
\]
where \( [f(x)]_y \) represents the score of the true label, thus quantifying the classifier's confidence in its prediction.

\begin{figure}[t]
\centering
\includegraphics[width=150mm]{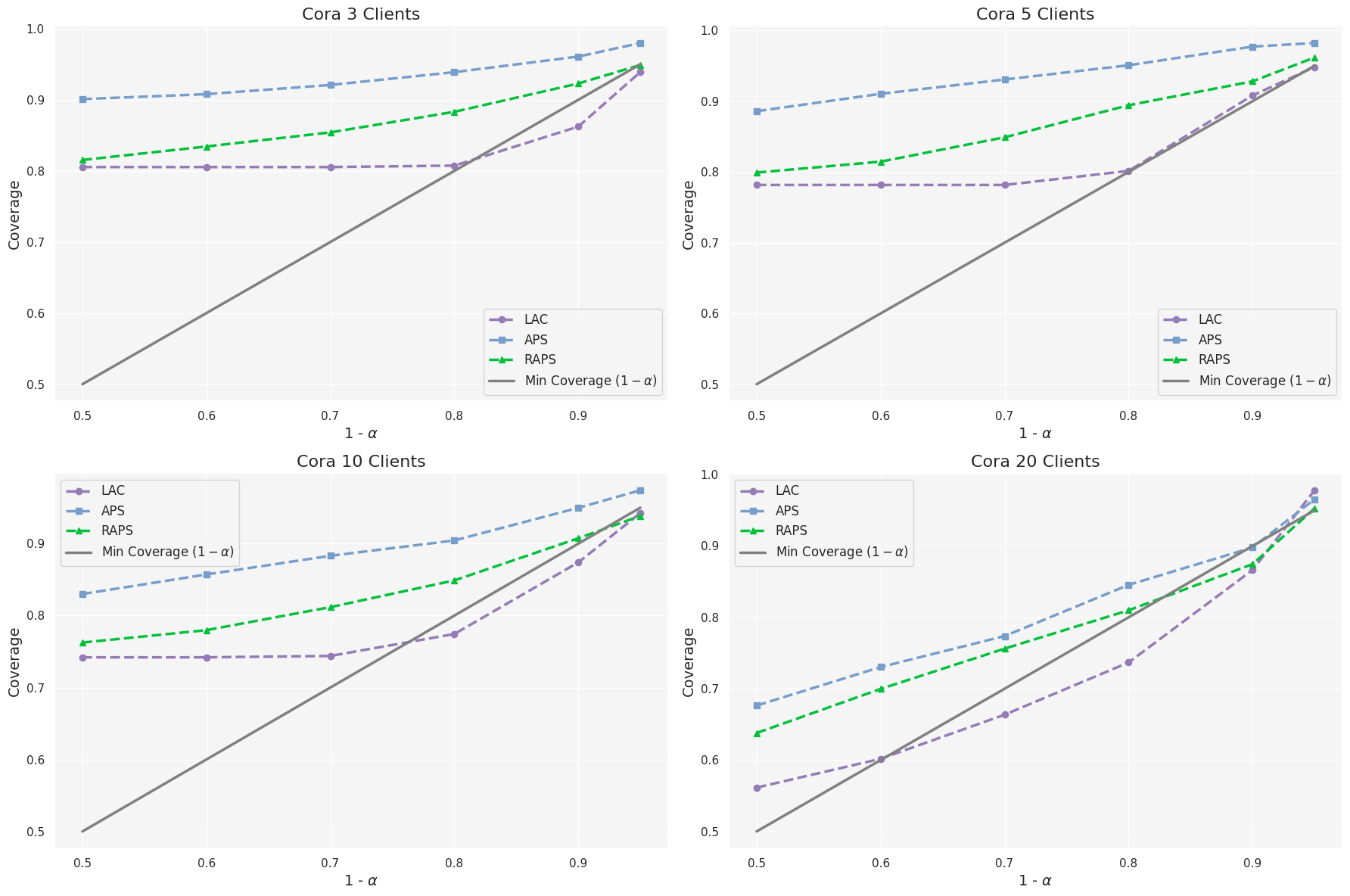}
\caption{\textbf{Coverage rates with different non-conformity scores for \texttt{Fed} model across varying \( K \) on the Cora dataset.}} \label{coverage}
\end{figure}

Table \ref{non-conformity} provides the comparison of CP set sizes when APS, RAPS, and LAC are employed as the non-conformity scores. While our proposed generative model improves the set size efficiency regardless of the non-conformity score used, LAC performs the best in most scenarios. However, when we examine the coverage rates, LAC is found to violate the coverage guarantee returning lower rates than the $1 - \alpha$ threshold as the number of clients increases as shown in Figure \ref{coverage}. 
 
\begin{table}[h]
\centering
\caption{CP set size comparison of non-conformity scores APS, RAPS and LAC on Cora dataset with partition number $K = 3, 5, 10$ and $20$. Set sizes are presented for $1 - \alpha = 0.95, 0.90$, and $0.80$ confidence levels. The corresponding std. are given with an averaged set size over 10 runs.}
\label{non-conformity}
\setlength{\tabcolsep}{6pt}
\begin{tabular}{@{}lccc ccc@{}}
\cmidrule(lr){2-7}
& \multicolumn{1}{c}{\textbf{APS}} & \multicolumn{1}{c}{\textbf{RAPS}} & \multicolumn{1}{c}{\textbf{LAC}} 
& \multicolumn{1}{c}{\textbf{APS}} & \multicolumn{1}{c}{\textbf{RAPS}} & \multicolumn{1}{c}{\textbf{LAC}} \\
\cmidrule(lr){2-4} \cmidrule(lr){5-7}
& \multicolumn{3}{c}{\textbf{$K=3$}} & \multicolumn{3}{c}{\textbf{$K=5$}} \\
\cmidrule(lr){2-4} \cmidrule(lr){5-7}
\texttt{Fed (0.95)}   & 4.31\std{0.02} & 2.57\std{0.02} & \textbf{1.79}\std{0.01}  & 4.94\std{0.02} & 2.97\std{0.01} & \textbf{2.59}\std{0.03}\\     
\texttt{Gen (0.95)}   & 4.25\std{0.02} & 2.22\std{0.01} & \textbf{1.58\std{0.02}}  & 5.09\std{0.02} & 2.82\std{0.02} & \textbf{2.53}\std{0.04}\\
\cmidrule(lr){2-4} \cmidrule(lr){5-7}
\texttt{Fed (0.90)}   & 3.34\std{0.03} & 1.85\std{0.01} & \textbf{1.19}\std{0.01}  & 4.14\std{0.03} & 2.33\std{0.02} & \textbf{1.64\std{0.02}}\\     
\texttt{Gen (0.90)}   & 3.34\std{0.02} & 1.69\std{0.02} & \textbf{1.12}\std{0.01}  & 4.10\std{0.02} & 2.12\std{0.03} & \textbf{1.61}\std{0.01}\\
\cmidrule(lr){2-4} \cmidrule(lr){5-7}
\texttt{Fed (0.80)}   & 2.45\std{0.01} & 1.36\std{0.01} & \textbf{1.01}\std{0.01}  & 2.95\std{0.01} & 1.63\std{0.01} & \textbf{1.04}\std{0.00}\\     
\texttt{Gen (0.80)}   & 2.51\std{0.03} & 1.27\std{0.02} & \textbf{1.00}\std{0.00}  & 2.98\std{0.05} & 1.52\std{0.02} & \textbf{1.04}\std{0.00}\\
\cmidrule(lr){2-7}

& \multicolumn{3}{c}{\textbf{$K=10$}} & \multicolumn{3}{c}{\textbf{$K=20$}} \\
\cmidrule(lr){2-4} \cmidrule(lr){5-7}
\texttt{Fed (0.95)}   & 5.02\std{0.02} & \textbf{3.50}\std{0.01} & 3.82\std{0.02}  & 5.79\std{0.02} & \textbf{5.16}\std{0.02} & 5.64\std{0.01}\\     
\texttt{Gen (0.95)}   & 4.86\std{0.02} & \textbf{3.39}\std{0.03} & 3.39\std{0.04}  & 5.40\std{0.02} & \textbf{4.92}\std{0.05} & 5.06\std{0.03}\\
\cmidrule(lr){2-4} \cmidrule(lr){5-7}
\texttt{Fed (0.90)}   & 4.32\std{0.02} & 2.61\std{0.01} & \textbf{2.06}\std{0.01}  & 4.13\std{0.01} & 3.78\std{0.01} & \textbf{3.37}\std{0.00}\\     
\texttt{Gen (0.90)}   & 3.98\std{0.01} & 2.55\std{0.02} & \textbf{2.00}\std{0.01}  & 3.90\std{0.04} & 3.55\std{0.03} & \textbf{3.05}\std{0.01}\\
\cmidrule(lr){2-4} \cmidrule(lr){5-7}
\texttt{Fed (0.80)}   & 2.93\std{0.03} & 1.79\std{0.01} & \textbf{1.19}\std{0.00}  & 3.17\std{0.03} & 2.92\std{0.01} & \textbf{2.27}\std{0.01}\\     
\texttt{Gen (0.80)}   & 2.92\std{0.02} & 1.73\std{0.01} & \textbf{1.14}\std{0.02}  & 2.88\std{0.03} & 2.50\std{0.01} & \textbf{1.73}\std{0.03}\\

\bottomrule
\end{tabular}
\end{table}

\section{Impact Quantile Averaging Methods}
\begin{figure}[t]
\centering
\includegraphics[width=150mm]{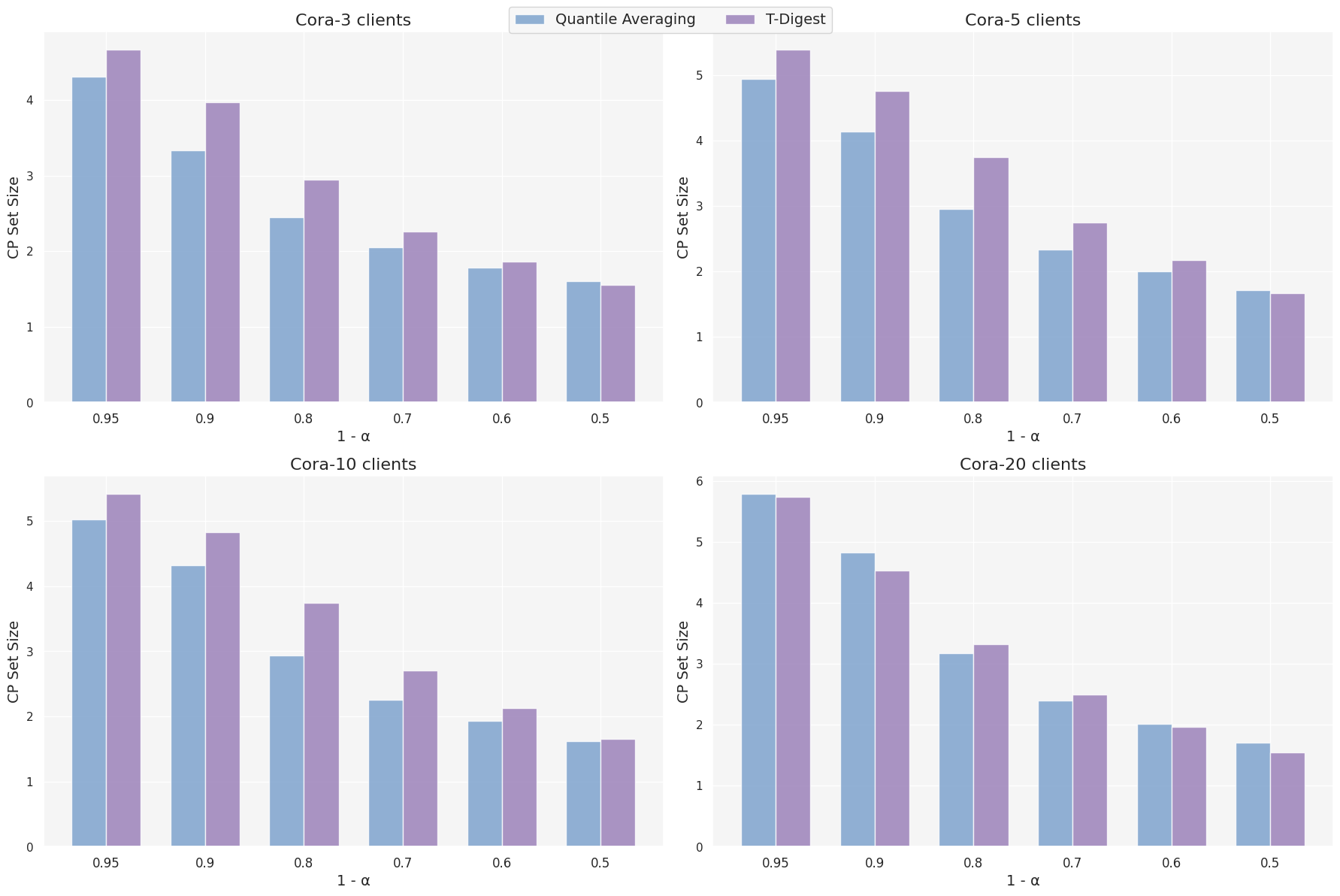}
\caption{\textbf{Comparison of T-Digest and quantile averaging methods by confidence level on Cora dataset.}} \label{digest}
\end{figure}

In this study, we evaluated the performance of two distributed quantile estimation methods, T-Digest \citep{dunning2021t} and quantile averaging \citep{luo2016quantiles}, with respect to their impact on conformal prediction set sizes. T-Digest is a probabilistic data structure optimized for the estimation of quantiles in extensive and distributed datasets, facilitating real-time analysis. Its mergeable nature enables effective aggregation of summaries across parallel, distributed systems, ensuring statistical efficiency and scalability. As shown in Figure \ref{digest}, T-Digest produces larger set sizes across various configurations of confidence levels and number of clients. We found quantile averaging is more effective at reducing model uncertainty.

\vfill

\clearpage


\bibliographystyle{apalike}
\bibliography{bibliography}

\end{document}